\documentclass[10pt,twocolumn,letterpaper]{article}

\usepackage{wacv}
\usepackage{times}
\usepackage{epsfig}
\usepackage{graphicx}
\usepackage{amsmath}
\usepackage{amssymb}
\usepackage{tabularx}
\usepackage{booktabs}

\usepackage{multirow}
\usepackage{makecell}
\usepackage{arydshln}
\usepackage{enumitem}
\usepackage[nocompress]{cite} 
\makeatletter
\newcommand\footnoteref[1]{\protected@xdef\@thefnmark{\ref{#1}}\@footnotemark}
\makeatother

\wacvfinalcopy 

\def\wacvPaperID{1241} 

\ifwacvfinal\fi
\begin{document}

\title{Hand-Priming in Object Localization for Assistive Egocentric Vision}

\author{Kyungjun Lee \hspace{2cm} Abhinav Shrivastava \hspace{2cm} Hernisa Kacorri \\
University of Maryland, College Park\\
{\tt\small kjlee@cs.umd.edu \hspace{1cm} abhinav@cs.umd.edu \hspace{1cm} hernisa@umd.edu}
}

\maketitle
\ifwacvfinal\thispagestyle{empty}\fi

\begin{abstract}
Egocentric vision holds great promises for increasing access to visual information and improving the quality of life for people with visual impairments, with object recognition being one of the daily challenges for this population. While we strive to improve recognition performance, it remains difficult to identify which object is of interest to the user; the object may not even be included in the frame due to challenges in camera aiming without visual feedback. Also, gaze information, commonly used to infer the area of interest in egocentric vision, is often not dependable. However, blind users often tend to include their hand either interacting with the object that they wish to recognize or simply placing it in proximity for better camera aiming. We propose localization models that leverage the presence of the hand as the contextual information for priming the center area of the object of interest. In our approach, hand segmentation is fed to either the entire localization network or its last convolutional layers. Using egocentric datasets from sighted and blind individuals, we show that the hand-priming achieves higher precision than other approaches, such as fine-tuning, multi-class, and multi-task learning, which also encode hand-object interactions in localization.

\end{abstract}


\section{Introduction}
\begin{figure}
    \centering
    \includegraphics[width=0.47\textwidth]{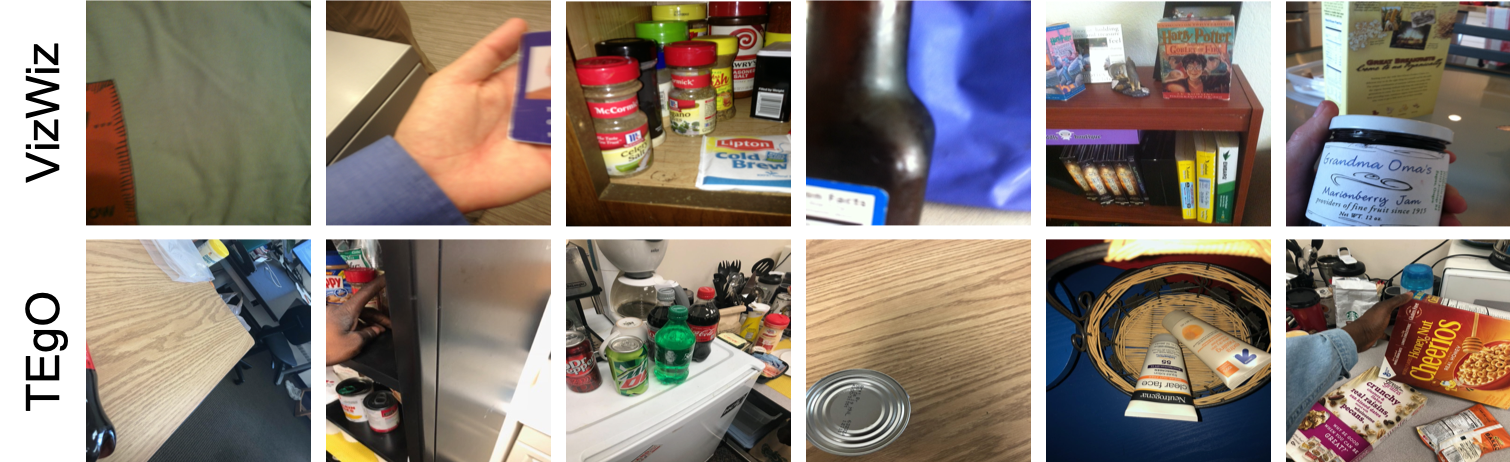}
    \caption{Examples of egocentric photos taken by people with visual impairments for object recognition in the crowdsourcing app, VizWiz~\cite{gurari2018vizwiz}, and a personalized fine-grained object recognition model in TEgO~\cite{lee2019hands}.  These examples illustrate the need for object-of-interest localization for better camera framing.}
    \label{fig:blind_examples}
\end{figure}

Computer vision holds a great promise for solving daily challenges that people with visual impairments face; one of the challenges is object recognition from egocentric vision~\cite{brady2013visual, bertasius2017unsupervised, damen2018scaling}. However, there is no guarantee that, without visual feedback, these users can aim the camera properly to indicate objects of interest in the frame. Consider the photos in Figure~\ref{fig:blind_examples} taken by people with visual impairments who tried identifying objects with the help of a sighted crowd on VizWiz~\cite{gurari2018vizwiz} or with their personalized object recognizer~\cite{lee2019hands}. Do all the images contain the object of interest? Do we know about which of the objects is the user inquiring?  Does the object show discriminative viewpoints? How can the user tell if the wrong object is being recognized given a cluttered background? These questions highlight the need for non-visual feedback that guides users to well-framed images of objects for the recognition task. While this is often achieved through a few iterations with sighted help in a crowdsourcing platform, it remains a challenging task for automated solutions, making recognition errors perceptible only through sight.  

In response, many non-visual feedback mechanisms have been proposed to help people with visual impairments take better quality photos for identifying objects~\cite{bigham2010vizwizlocateit, white2010easysnap, jayant2011supporting}. The most recent work in this direction explores the utility of proprioception~\cite{sherrington1907proprio, proske2012proprioceptive}, the perception of body and limb position, in the context of object recognition. It shows that many people with visual impairments naturally tend to hold or place their hand close to the object of interest for better camera framing~\cite{lee2019hands}. This finding is further supported by prior evidence on the ability of blind people to guide hand orientation and to make rapid corrections through proprioception~\cite{gosselin2009evidence}. While the presence and shape of the hand are shown to be helpful for estimating object centers and providing real-time non-visual feedback, prior work also shows that people with visual impairments are susceptible to false positives as they have no means to verify whether the localization model is correct --- they trust the feedback even though they know that it can be wrong~\cite{lee2019revisiting}. 
 
Towards a feedback mechanism with lower false positives for object-of-interest localization, we present a computational model that builds on prior work in contextual priming~\cite{shrivastava2016contextual}. In our approach, hand segmentation is used to provide contextual information to guide the localization of the object of interest in static images.
The intuition is that the hand segmentation can help capture semantic relations between the hand and the object of interest, such as the relationship between the hand position and the object position and the relationship between the hand pose and the object size~\cite{biederman2017semantics}.  Such relations will essentially guide the localization model to pay attention to certain regions near the hand in the image. We explore the potential of context priming by infusing hand segmentation to: (i) the entire localization network or (ii) the last two convolutional layers only.
As in~\cite{shrivastava2016contextual}, the first approach is designed to impart the hand features in all layers. The second approach is based on the observation that later layers capture more descriptive features~\cite{zeiler2014visualizing}, such as hand features captured in a later convolutional layer~\cite{ma2016going}.

We evaluate these approaches on three egocentric datasets, GTEA~\cite{fathi2011learning}, GTEA Gaze+~\cite{fathi2012learning}, and TEgO~\cite{lee2019hands}, by comparing the performances of our methods with those of other methods: a naive no-hand object localization model, a object localization model that is fine-tuned from the hand segmentation model~\cite{ma2016going, lee2019hands}, and alternative approaches that frame object localization as a multi-class or multi-task problem. Our evaluation finds that hand-priming can contribute to better localization of the object of interest in terms of false positives, especially when incorporated in the later convolutional layers. While there are limited annotated data from people with visual impairments, the result on images from a blind individual seems to be consistent.

To the best of our knowledge, this is the first work to provide empirical results from different approaches on object-of-interest localization for an assistive egocentric-vision task that can benefit people with visual impairments. A unique challenge in this task is that gaze information, commonly used to infer the individuals' area of interest in egocentric vision~\cite{fathi2012learning, li2015delving, zuo2018gaze}, is not available for people with visual impairments, but hand--object interactions can be leveraged to fill this gap through context priming. 

\section{Related Work}
Our work draws upon prior work on egocentric object localization.  We discuss previous attempts to understand hand--object interactions in egocentric vision and focus on models that are augmented by contextual feedback, which inspire our hand-priming approach for object localization. Then, we shift our discussion to assistive technologies that employ computer vision models to understand egocentric data collected by people with visual impairments.

\subsection{Egocentric Interactions}

Interactions between hands and objects in egocentric vision have been explored for various tasks: from gaze estimation to human behavior understanding.
The estimation of hand pose and shape can serve as a cue to understanding users' intentions and thus has been the focus of research for both the computer vision and human-computer interaction communities~\cite{rogez20143d, choi2017robust, lee2019hands, tekin2019h+}.

Traditionally, the egocentric hand information has been utilized in activity recognition~\cite{surie2007activity, fathi2011learning, pirsiavash2012detecting, ma2016going}.  Prior work focuses on the interactions between hands and objects due to that objects can possess a clue to users' activities~\cite{surie2007activity, pirsiavash2012detecting, li2015delving, ma2016going}.  Using two-stream convolutional neural networks, Ma and Kris~\cite{ma2016going} try to localize and recognize an object of interest according to the hand pose and location in egocentric vision to recognize users' activities.  This prior work suggests localizing the center area of the object of interest, instead of its exact center coordinates. Our method also follows this approach for object localization.

Some prior work assumes that some implicit visual attention, such as gaze information, has been implicitly encoded in input data; hence, there is prior work that tries to estimate gaze points in egocentric data~\cite{li2013learning, li2018eye, huang2018predicting, tavakoli2019digging} or uses this visual attention to accomplish other tasks~\cite{fathi2012learning, li2015delving, zuo2018gaze}.
Such visual attention, however, may not be encoded in data from different populations --- for instance, blind people.
For this issue, recent prior work has proposed to use more explicit visual attention, such as fingers pointing to the item of interest, to understand items of interest of blind people in complex visual scenes~\cite{guo2018investigating}.
In addition, even for people with visual impairments, hands have been found to be the explicit cue to objects of their interest~\cite{kacorri2017people, lee2019hands}.
Based on these prior observations, our work focuses on interactions between egocentric hands and objects of interest to learn the relationship between the hands and objects of interest.

Hand--object interactions have also been explored for other tasks.
Tekin \etal~\cite{tekin2019h+} propose a single neural network model that outputs 3D hand and object poses from RGB images and recognizes objects and users' actions.
Cai \etal~\cite{cai2016understanding} suggest to model the relationship between hand poses and object attributes, as those data can provide complementary information to each other. With the model that understands the hand--object relationship, this prior work further tries to recognize users' actions.
As our model localizes an object of interest based on the hand--object interactions, we see its potential usage in other applications, such as recognition of objects and actions.

\subsection{Context Reinforcement in Vision}
Prior work in cognitive science has shown that the human visual system tries to use ``context'' to pay attention only to our interest~\cite{tulving1990priming, wig2005reductions, meng2014neural}.
The use of the context information has also been discussed from the perspective of computer vision~\cite{palmer1975effects, oliva2007role, divvala2009empirical}. Inspired by these observations, researchers in computer vision have introduced various approaches that exploit contextual information~\cite{torralba2003context, murphy2004using, rabinovich2007objects, yao2012describing, mottaghi2014role}.

In the deep learning era, researchers have also shown that using contextual information, such as instance segmentation, helps to improve the performance of object detection and recognition~\cite{zhu2015segdeepm, shrivastava2016contextual, li2017fully, he2017mask}.
FCIS~\cite{li2017fully} and Mask R-CNN~\cite{he2017mask} are designed to learn instance segmentation, which is then used for object classification.
On the other hand, Shrivastava and Gupta~\cite{shrivastava2016contextual} proposed a slightly different approach.  In their model, a segmentation module is used to provide contextual feedback in the object detection model (Faster R-CNN~\cite{ren2015faster}); \ie, the segmentation output is appended to the object detection model.
Built upon this contextual priming and feedback approach, our approach takes into account the contextual information (\ie, the hand information) for object localization in egocentric vision.

\subsection{Egocentric Vision in Assistive Technologies}
Assistive technologies, especially for people with visual impairments, have employed computer vision algorithms to help them access visual information that surrounds them.
As people with visual impairments use their smartphones or wearable cameras to take photos of their target, the captured data are in egocentric vision; thus, assistive technologies have focused on computer vision models that can understand the egocentric vision.
For instance, researchers in assistive technologies have proposed various approaches that can help visually impaired people capture their surroundings and access to that information.
Such technologies, called \textit{blind photography}, have been developed to simply capture scenes~\cite{vazquez2012helping, vazquez2014assisted}, and recognize objects~\cite{bigham2010vizwizlocateit, white2010easysnap, jayant2011supporting, zhong2013real} or people (\eg, family members and friends)~\cite{white2010easysnap, balata2015blindcamera, zhao2018face}.
In particular, prior work on blind photography reported that people with visual impairments are vulnerable to false-positive object localization~\cite{lee2019revisiting} as there are no means for these users to check the localization output. In this paper, considering this prior finding, we focus on reducing false positives in object localization and evaluating the object localization models with the precision metric as this measurement captures false-positive estimations.


\section{Methods}

\begin{figure}[!t]
    \centering
    \includegraphics[width=0.48\textwidth]{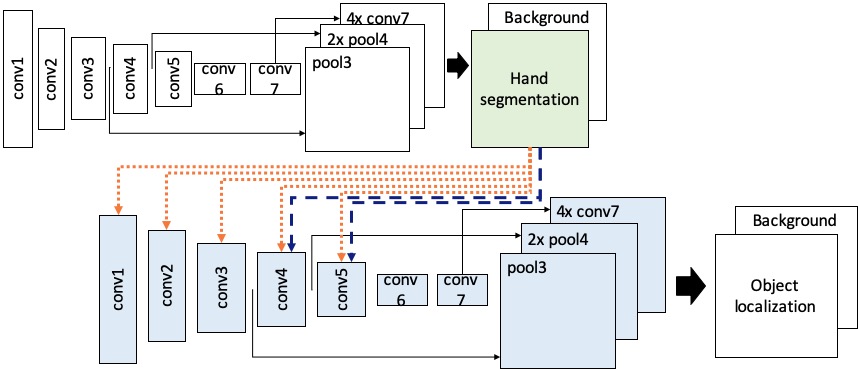}
    \caption{The architecture of our hand-primed object localization model. It consists of two neural networks: one (above) for hand segmentation and another (below) for object localization. The hand segmentation output is infused into the object localization model in two ways: \textit{HPAll} (orange dotted line) and \textit{HPLate} (blue dashed line).}
    \label{fig:arch}
\end{figure}

We introduce a hand-primed object localization model that is built upon the prior work on contextual priming and feedback~\cite{shrivastava2016contextual}. The objective of our model is to localize an object of interest pertaining to the hand information in egocentric vision.

\subsection{Hand-Primed Object Localization}

We build an object localization model that is reinforced by a hand segmentation output.
Our hand-primed object localization model consists of two models: a hand segmentation network and an object localization network.
For both of the models, we employ the FCN-8s architecture~\cite{long2015fully}.  Using the pixel-level classification model, our object localization model tries to estimate which pixels belong to the center area of an object of interest as proposed in~\cite{ma2016going}; the one proposed in this prior work is used as the baseline model that uses the hand information for object localization.

As depicted in Figure~\ref{fig:arch}, the hand segmentation output is appended to several layers in the object localization network to prime the model to use the hand information for object localization.
We propose two different ways of priming the object localization model to use the hand information: \textit{HandPrimedAll (HPAll)} and \textit{HandPrimedLate (HPLate)}.
Assuming that providing this segmentation to all layers would make a model primed to use that feedback for its tasks~\cite{shrivastava2016contextual}, we design the HPAll model in which the segmentation output is infused to all the convolutional layers (\textit{conv1--conv5}).
On the other hand, in the HPLate model, the hand segmentation output is infused into the later convolutional layers (\textit{conv4, conv5}).  This approach is inspired by the observations that the later convolutional layers capture the more descriptive features~\cite{zeiler2014visualizing}, and the hand features are found to be captured in a later convolutional layer (\textit{conv5}) of the prior object localization network~\cite{ma2016going}.

The hand segmentation model has two output layers to segment hands and a background in an image; \ie, the model performs a per-pixel classification to determine which pixels belong to the hand or the background.
For object localization, our model does not estimate the exact center and bounding box of an object of interest.  Instead, the model infers the center area of a target object.  Having two output layers, our hand-primed object localization model determines which pixels belong to the center area of an object of interest or the non-center area (background).
Prior work in pose estimation~\cite{pfister2015flowing} and egocentric activity recognition~\cite{ma2016going} has highlighted benefits of estimating areas of interest over the exact coordinate estimation.

Our model expects egocentric images that are obtained by users taking photography or by extracting salient frames from videos recorded using a wearable camera on the head, the chest, or eyeglasses.

\subsection{Implementation Details}
As the object localization output is dependent on the hand segmentation output, we separately trained these two networks.  First, we trained the hand segmentation network.  Then, while freezing the weights of the hand segmentation network, we trained the object localization model.

Adam~\cite{kingma2014adam} was used to train our hand segmentation and object localization networks.
Following are hyper-parameters that we set for training:
(hand segmentation) 10,000 training steps, 0.00001 learning rate, 16 batch size, and $10^{-9}$ Adam's epsilon;
(object localization) 20,000 training steps, 0.00001 learning rate, 8 batch size, and $10^{-9}$ Adam's epsilon.
In both of the models, we initialized the weights of the first five convolutional layers (\textit{conv1--conv5}) with those of the VGG-16 network model~\cite{simonyan2014very} pre-trained on ImageNet~\cite{deng2009imagenet}.

\begin{figure}[!t]
    \centering
    \includegraphics[width=0.37\textwidth]{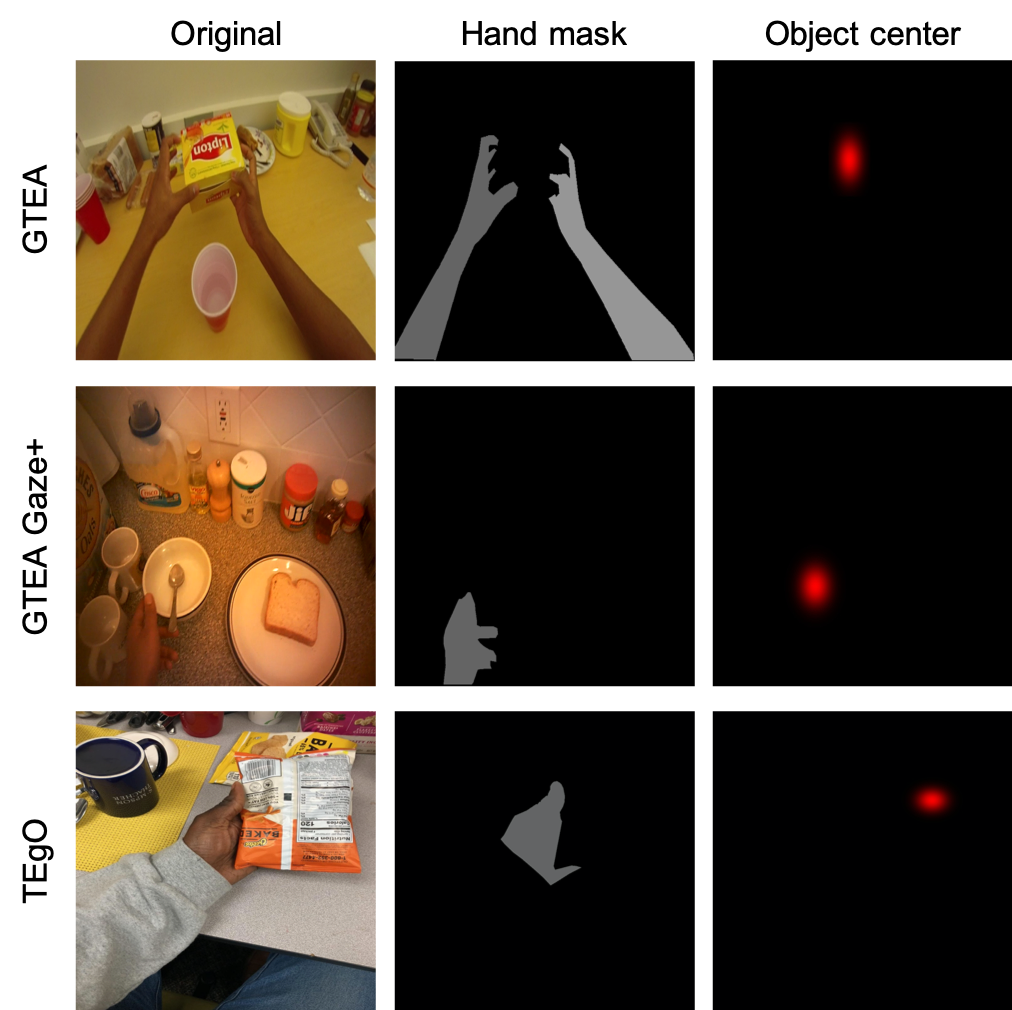}
    \caption{Input data for training models: an original image, its hand mask, and its object center annotation.}
    \label{fig:model_input}
\end{figure}

\begin{description}[leftmargin=0cm]
    \item[Training:] Using four egocentric datasets including EgoHands~\cite{bambach2015lending}, GTEA~\cite{fathi2011learning}, GTEA Gaze+~\cite{fathi2012learning}, and TEgO~\cite{lee2019hands}, we first trained our hand segmentation model. Provided with a set of original images and those hand masks as shown in Figure~\ref{fig:model_input}, our hand segmentation model learns hand features from the input data during training.
    To train our object localization model, a set of original images and those object center annotation data were used; Figure~\ref{fig:model_input} shows the examples. The center of an object of interest in each image was annotated with a Gaussian heatmap blob. The object localization model was taught to estimate the center area of a target object interacted with the hand; \ie, we trained the model to learn the relationship between the hand pose and object location. GTEA, GTEA Gaze+, and TEgO were used to train and test the object localization model. Note that we did not use the EgoHands dataset for object localization as this dataset mostly contains hand movements in board games, such as chess, card games, and Jenga.
    As both of the hand segmentation and object localization models predict a class per pixel, the cross-entropy loss function was used to train both of the networks.
    
    \item[Inference:] At test time for object localization, we only considered per-pixel classification outputs which confidence scores were higher than 0.5 in the object localization output layer. When there was more than one cluster of the estimated center area, the biggest cluster was chosen to be the estimated center area.
    
\end{description}

\section{Experiments}
For evaluation, using egocentric hand--object datasets, we compared our approach with several other approaches of using the hand information to localize an object of interest.
As the localization outputs and these ground-truths were represented as an area, $mIoU$ (mean intersection over union) and the standard COCO metrics, including $AP$ (averaged over IoU thresholds), $AP_{50}$, and $AP_{75}$, were used to report the performance of each model. $AP$ is an appropriate metric for our evaluation in that it inherently captures false-positive predictions~\cite{davis2006relationship}, which are a critical factor in object localization.
It also measures the capability of a model to be employed in assistive technologies for people with visual impairments, such as blind photography, since users of such systems are prone to false-positives~\cite{lee2019revisiting}.

\subsection{Datasets}
All the datasets (GTEA, GTEA Gaze+, TEgO) already provide hand segmentation data for their original images, but only the TEgO dataset provides the annotation data for object location; they used a Gaussian heatmap blob to indicate the center of an object of interest, which was also used for egocentric activity recognition~\cite{ma2016going}.  For the other datasets (GTEA and GTEA Gaze+), we used the object localization annotation provided by Lee and Kacorri~\cite{lee2019hands}; they shared their manual annotation data for these datasets on the TEgO website.
Following is the number of images used from each dataset: GTEA (663), GTEA Gaze+ (1,115), and TEgO (5,758). Note that we excluded images without hands from the TEgO dataset in our evaluation.  We randomly selected samples from each dataset and applied the following dataset splitting to each dataset: 80\% for training, 10\% for validation, and 10\% for testing.

The GTEA datasets include images only collected by sighted people, but the TEgO dataset contains data from a blind person.
The evaluation with this TEgO data would gauge the compatibility of each localization model with assistive technologies for people with visual impairments, such as blind photography or object recognition.


\subsection{Comparison Models}
There were four different models with which we used to compare our approach. Similar to our model, all the comparison models were also built upon the FCN-8s model. As in our method, the Adam optimizer was used for training.

\begin{description}[leftmargin=0cm]
    \setlength\itemsep{0em}
    \item [NoHand:] This model was trained\footnote{\label{note:hyperparams1}10,000 training steps, 0.00001 learning rate, 16 batch size.} for object localization without explicitly being trained on hand segmentation data; \ie, the object center annotation was only used for training. 
    \item [Finetune:] This is the baseline model that uses hand features for object localization, which was introduced by Ma \etal~\cite{ma2016going}. To implement this model, we fine-tuned\footnoteref{note:hyperparams1} our hand segmentation model to the object localization problem for each dataset. During the fine-tuning, we froze the weights of the first five convolutional layers (\textit{conv1--conv5}) as in the model proposed by the prior work; they observed that hand features were captured in the conv5 layer.
    \item [MultiClass/MultiClass-2x:] In this model, learning the hand and the object center was considered as a multi-class problem. This multi-class model estimates the class of each pixel; \ie, it determines to which class among \textit{background}, \textit{hand}, and \textit{object center area} each pixel belongs. In addition, we implemented another multi-class model, \textit{MultiClass-2x}, in which we set twice more weight on the loss in object localization. For both of the models, the weights were also initialized with those of our object hand segmentation model before training\footnoteref{note:hyperparams1}.
    \item [MultiTask/MultiTask-2x:] Considering hand segmentation and object localization as two different tasks in one model, we employed the hard parameter sharing for this multi-task learning~\cite{Caruana1993MultitaskLA}. This model shares the first five convolutional layers (\textit{conv1--conv5}) and has two separate branches --- one branch for the hand segmentation and another for the object localization.   This model is called \textit{MultiTask}. Moreover, as in the MultiClass models, we also created another multi-task model in which the loss in object localization model was computed to be twice more important than the loss in hand segmentation. We call this model \textit{MultiTask-2x}.  Before training\footnote{\label{note:hyperparams2}20,000 training steps, 0.00001 learning rate, 8 batch size.} these models, we also initialized the weights of the hand segmentation part of these models with those of our hand segmentation model; \ie, the weights of the shared convolutional layers and the separate layers for the hand segmentation part were initialized with those of our hand segmentation model.
\end{description}



\subsection{Hand Segmentation}

\begin{table}[!b]
    \centering
    \caption{Quantitative analysis of the hand segmentation models. The average interaction of union ($mIoU$) and the COCO standard metrics ($AP$, $AP_{50}$, $AP_{75}$) are used.}
    \resizebox{\columnwidth}{!}{
    \begin{tabular}{@{}llcccc@{}}
        \toprule
        Dataset & Model & $mIoU$ & $AP$ & $AP_{50}$ & $AP_{75}$\\
        \midrule
        \multirow{6}{*}{GTEA} & NoHand & N/A & N/A & N/A & N/A \\
                              & MultiClass & 0.82 & 0.68 & 1.0 & 0.85 \\
                              & MultiClass-2x & 0.77 & 0.58 & 0.98 & 0.63 \\
                              & MultiTask & 0.92 & 0.88 & 1.0 & 0.98 \\
                              & MultiTask-2x & 0.92 & 0.88 & 1.0 & 0.98 \\
                              & Finetune,\textit{HPAll/Late} & 0.9 & 0.85 & 1.0 & 0.98 \\
        \midrule
        \multirow{6}{*}{\makecell{GTEA\\Gaze+}} & NoHand & N/A & N/A & N/A & N/A \\
                              & MultiClass & 0.88 & 0.81 & 0.98 & 0.93 \\
                              & MultiClass-2x & 0.83 & 0.72 & 0.97 & 0.83 \\
                              & MultiTask & 0.92 & 0.88 & 1.0 & 0.96 \\
                              & MultiTask-2x & 0.91 & 0.86 & 1.0 & 0.95 \\
                              & Finetune,\textit{HPAll/Late} & 0.91 & 0.87 & 0.99 & 0.97 \\
        \midrule
        \multirow{6}{*}{TEgO} & NoHand & N/A & N/A & N/A & N/A \\
                              & MultiClass & 0.92 & 0.9 & 0.99 & 0.98 \\
                              & MultiClass-2x & 0.89 & 0.83 & 0.98 & 0.95 \\
                              & MultiTask & 0.92 & 0.89 & 0.98 & 0.97 \\
                              & MultiTask-2x & 0.92 & 0.91 & 0.98 & 0.97 \\
                              & Finetune,\textit{HPAll/Late} & 0.92 & 0.91 & 0.98 & 0.98\\
        \bottomrule
    \end{tabular}
    }
    \label{tab:hand_quant_analysis}
\end{table}

We first evaluate the hand segmentation performance of the models as our work focuses on this information to localize an object of interest in egocentric vision.
A threshold of $0.5$ was used to determine the per-pixel classification on the hand segmentation output layer.
We measured the hand segmentation performance of \textit{Finetune} before fine-tuning it to object localization; thus, three models (\textit{Finetune}, \textit{HPAll}, and \textit{HPLate}) ended up having the same hand segmentation model.
Note that the hand segmentation part for the multi-class (\textit{MultiClass}, \textit{MultiClass-2x}) and multi-task (\textit{MultiTask}, \textit{MultiTask-2x}) models was initialized with the weights of our hand segmentation model and then fine-tuned to each dataset again.

\begin{figure}[!t]
    \centering
    \includegraphics[width=0.47\textwidth]{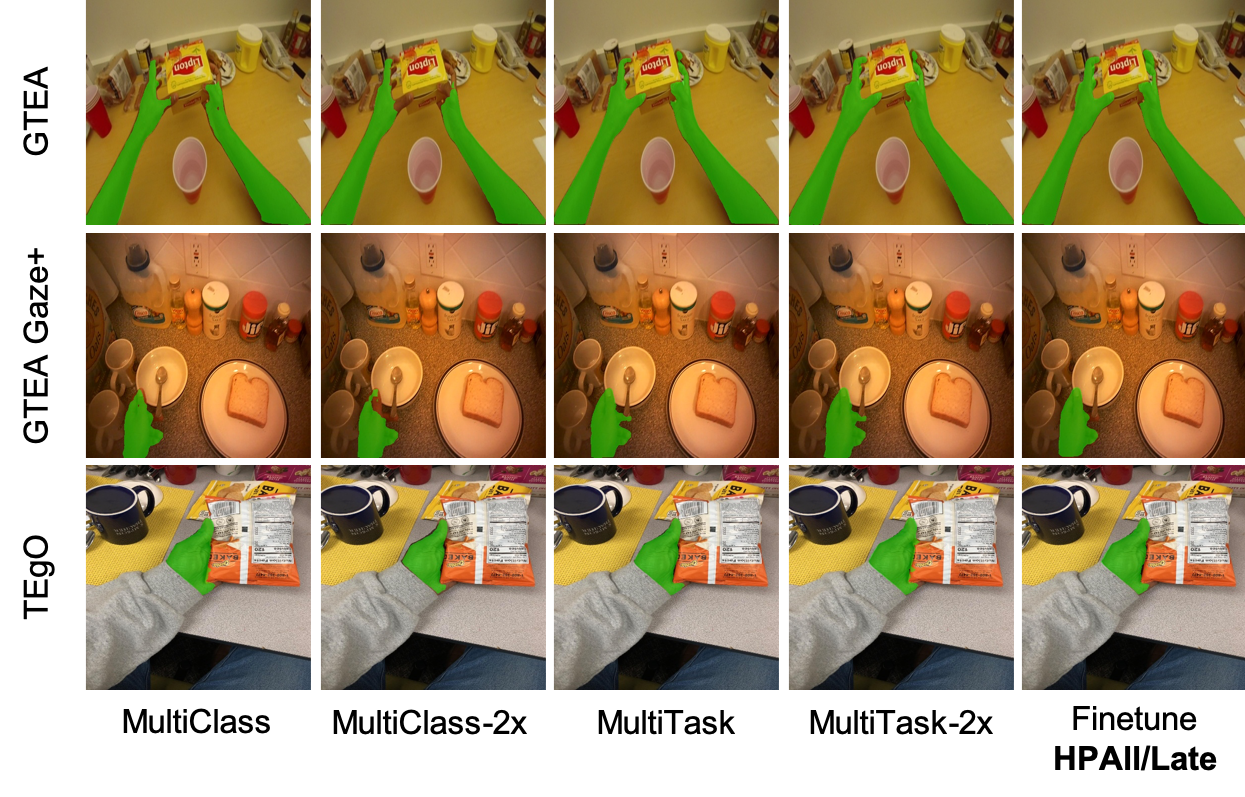}
    \caption{Hand segmentation outputs of the comparison models and our approaches (\textit{HPAll}, \textit{HPLate}).  The hand segmentation is overlaid with the green color. All the models except for the multi-class models appropriately segmented whole hand(s) from the testing examples; the multi-class models failed to segment fingers from the examples from GTEA and GTEA Gaze+.}
    \label{fig:hand_output}
\end{figure}

\begin{figure*}[!t]
    \centering
    \includegraphics[width=0.87\textwidth]{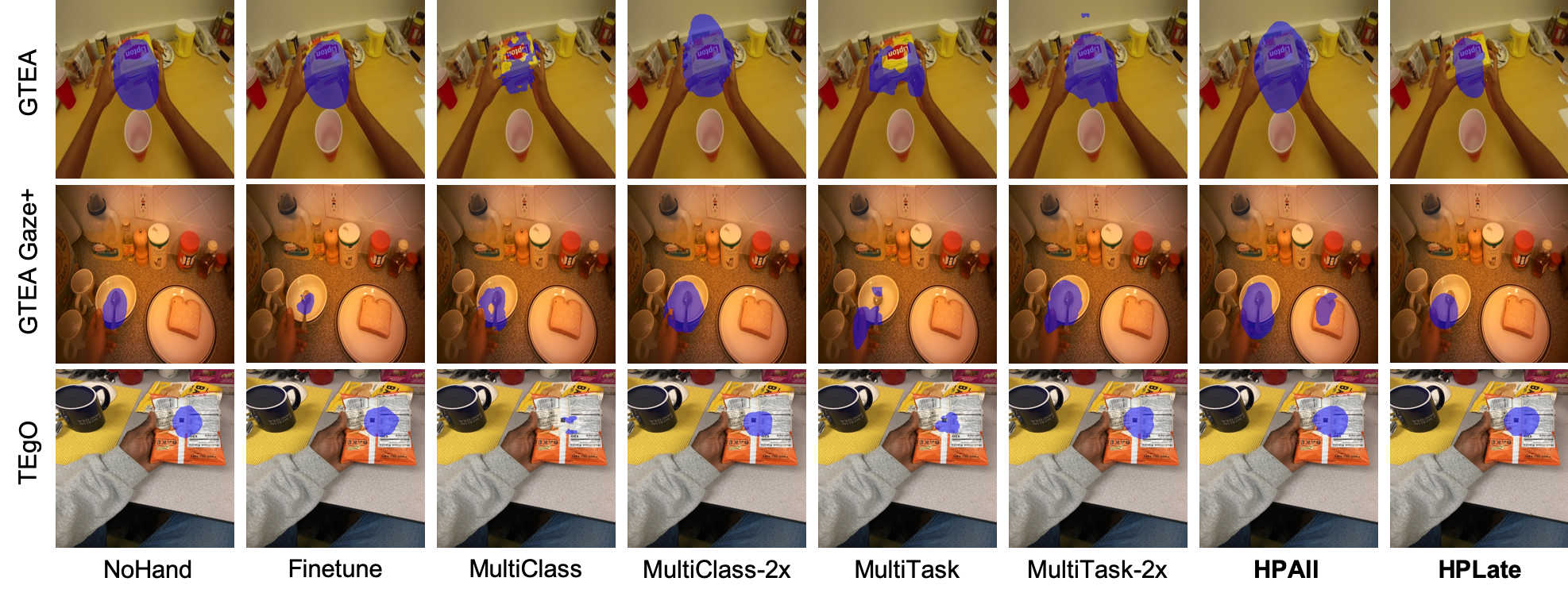}
    \caption{Object localization outputs of the comparison models and our approaches (\textit{HPAll}, \textit{HPLate}).  The object localization output is overlaid with the blue color on the testing examples; the threshold for the localization output was set to 0.5.}
    \label{fig:model_output}
\end{figure*}

\begin{description}[leftmargin=0cm]
    \item[Quantitative analysis:] We report the quantitative analysis of the hand segmentation performance of each model in Table~\ref{tab:hand_quant_analysis}. Note that \textit{NoHand} was not evaluated for hand segmentation since this model was trained only for object localization. As described in the table, our hand segmentation model and the multi-task models achieved equal to or higher than 0.9 in mIoU and 0.85 in AP in all of the three datasets. On the other hand, the hand segmentation performances of the multi-class models were not competitive to the other models including our hand segmentation model. In particular, although the performance of the localization-focused multi-task model (\textit{MultiTask-2x}) showed the similar performance to that of the naive multi-task model (\textit{MultiTask}), we observed the performance degradation in the localization-focused multi-class model (\textit{MultiClass-2x}) when comparing its performance to that of the naive multi-class model (\textit{MultiClass}) in all of the datasets.
    
    \item[Qualitative analysis:] The hand segmentation outputs of all the models (except for \textit{NoHand}) are visualized with the green color in Figure~\ref{fig:hand_output}. As in the quantitative analysis, our visual inspection informs that all the models were able to segment hand(s) from testing images, but the multi-class models lack detailed segmentation.  In general, the multi-class models segmented a large part of hand(s) from the testing images, but sometimes were unable to provide fine-grained hand segmentation. In the figure, we observed that the multi-class models were unable to segment fingers from the testing examples from GTEA and GTEA Gaze+.
\end{description}

\subsection{Object Localization}

As in the hand segmentation analysis, we evaluate the object localization performance, quantitatively and qualitatively.
A threshold of $0.5$ was used to determine whether per-pixel estimations belong to the center area of an object.

\begin{table}[!b]
    \centering
    \caption{Quantitative analysis of the object localization models. The average interaction of union ($mIoU$) and the COCO standard metrics ($AP$, $AP_{50}$, $AP_{75}$) are used.}
    \resizebox{\columnwidth}{!}{
    \begin{tabular}{@{}llcccc@{}}
        \toprule
        Dataset & Model & $mIoU$ & $AP$ & $AP_{50}$ & $AP_{75}$ \\
        \midrule
        \multirow{8}{*}{GTEA} & NoHand & 0.69 & 0.48 & 0.82 & 0.51 \\
                              & Finetune & 0.68 & 0.47 & 0.80 & 0.46 \\
                              & MultiClass & 0.26 & 0.01 & 0.06 & 0 \\
                              & MultiClass-2x & 0.66 & 0.48 & 0.80 & 0.51 \\
                              & MultiTask & 0.42 & 0.10 & 0.42 & 0 \\
                              & MultiTask-2x & 0.67 & 0.48 & 0.80 & 0.52 \\
                              & \textit{HPAll} & 0.70 & 0.51 & 0.77 & \textbf{0.58} \\
                              & \textit{HPLate} & \textbf{0.73} & \textbf{0.55} & \textbf{0.85} & 0.57 \\
        \midrule
        \multirow{8}{*}{\makecell{GTEA\\Gaze+}} & NoHand & 0.49 & 0.23 & 0.53 & 0.20 \\
                              & Finetune & 0.43 & 0.20 & 0.45 & 0.12 \\
                              & MultiClass & 0.24 & 0.01 & 0.07 & 0 \\
                              & MultiClass-2x & \textbf{0.60} & \textbf{0.33} & \textbf{0.74} & 0.24 \\
                              & MultiTask & 0.31 & 0.05 & 0.23 & 0 \\
                              & MultiTask-2x & 0.52 & 0.24 & 0.68 & 0.11 \\
                              & \textit{HPAll} & 0.50 & 0.25 & 0.55 & 0.20 \\
                              & \textit{HPLate} & 0.55 & 0.30 & 0.70 & \textbf{0.25} \\
        \midrule
        \multirow{8}{*}{TEgO} & NoHand & 0.69 & 0.48 & 0.86 & 0.47 \\
                              & Finetune & 0.70 & 0.49 & 0.88 & 0.51 \\
                              & MultiClass & 0.24 & 0.01 & 0.08 & 0 \\
                              & MultiClass-2x & 0.70 & 0.46 & 0.92 & 0.41 \\
                              & MultiTask & 0.37 & 0.02 & 0.14 & 0 \\
                              & MultiTask-2x & 0.72 & 0.52 & \textbf{0.94} & 0.52 \\
                              & \textit{HPAll} & 0.70 & 0.48 & 0.90 & 0.44 \\
                              & \textit{HPLate} & \textbf{0.74} & \textbf{0.55} & 0.93 & \textbf{0.59} \\
        \bottomrule
    \end{tabular}
    }
    \label{tab:local_quant_analysis}
\end{table}

\begin{description}[leftmargin=0cm]
    \item[Quantitative analysis:] Table~\ref{tab:local_quant_analysis} shows the object localization performances of our hand-primed object localization models (\textit{HPAll}, \textit{HPLate}) and the comparison models. In the GTEA and TEgO datasets, our model (\textit{HPLate}) outperformed the comparison models in almost all of the metrics.  In the GTEA Gaze+ dataset, our HPLate model showed the best performance in $AP_{75}$ and comparative performances to those of \textit{MultiClass-2x} in the other metrics, such as $mIoU$, $AP$, and $AP_{50}$.
    
    Between our hand-primed object localization models, \textit{HPLate} generally showed better performance than \textit{HPAll} in all of the datasets, which indicates that modulation of hand features in higher layers helps capture the contextual relationship between hands and the object of interest more than modulating hand features in all layers. Compared with the baseline model (\textit{Finetune}), our model (\textit{HPLate}) achieved the performance improvement by 17\% (from 0.47 to 0.55 AP), 50\% (from 0.2 to 0.3 AP), and 12\% (from 0.49 to 0.55 AP) in GTEA, GTEA Gaze+, and TEgo datasets, respectively. This result demonstrates that the explicit use (infusing) of the hand segmentation in object localization is more helpful for localizing an object of interest than fine-tuning the fixed features of the hand model to a different problem, object localization.
    Furthermore, in comparison to \textit{NoHand} that did not have the explicit training for hand information, our model (\textit{HPLate}) showed better object localization performance in all the three datasets. Based on these observations, it seems that having explicit hand features in the last convolutional layers (\eg, \textit{conv4} and \textit{conv5}) helps to prime the object localization model to focus on necessary features, such as hand pose and location.
    
    \item[Qualitative analysis:] Figure~\ref{fig:model_output} shows that our hand-priming models (\textit{HPAll}, \textit{HPLate}) well predicted the center area of an object associated with hand(s). As described in the quantitative analysis, \textit{HPLate} generated more precise localization outputs than \textit{HPAll}; the testing examples depict that \textit{HPAll} tended to output a larger center area than \textit{HPLate} and sometimes estimated more than one center area as shown in the GTEA Gaze+ example.
    \textit{MultiClass} and \textit{MultiTask} were unable to estimate the center area of an object of interest successfully, while the localization-focused version of these models (\textit{MultiClass-2x}, \textit{MultiTask-2x}) improved the performance of object localization.

\begin{figure}[!t]
    \centering
    \includegraphics[width=0.47\textwidth]{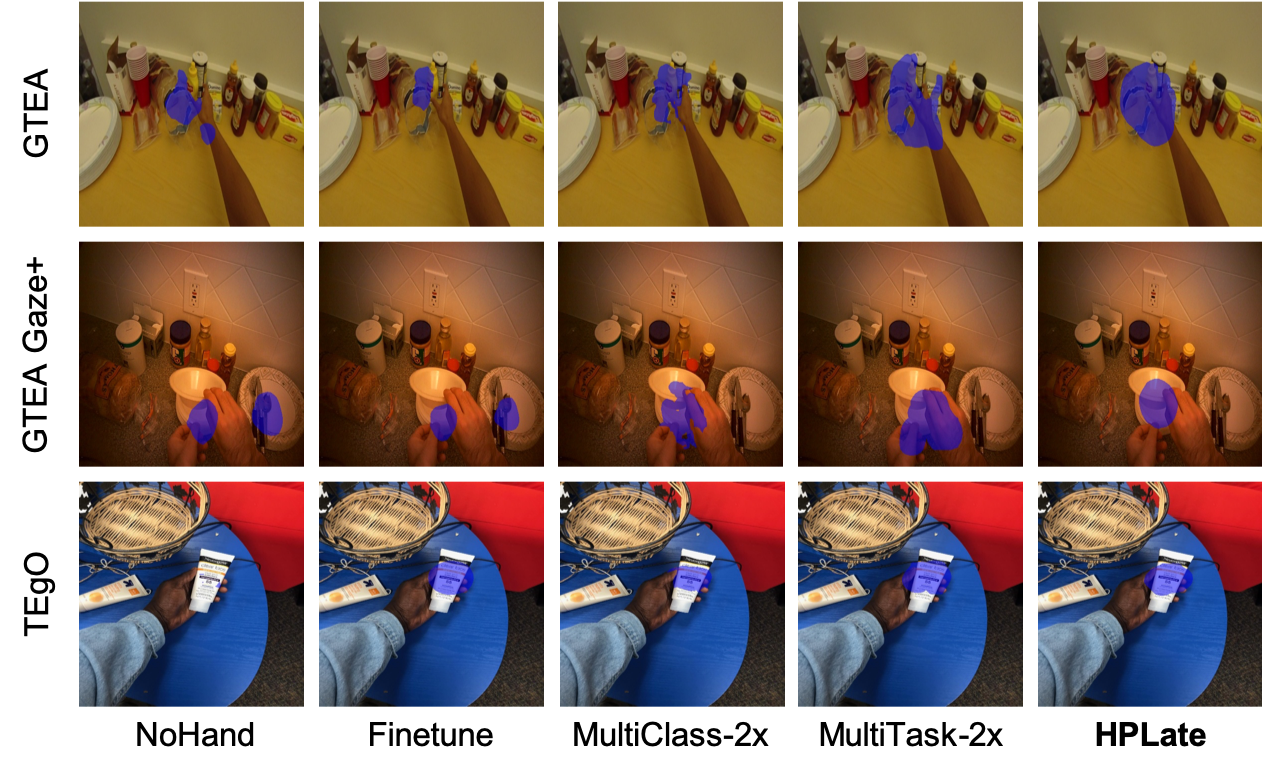}
    \caption{Object localization outputs of the models on other testing examples. In comparisons with \textit{NoHand}, our method (\textit{HPLate}) shows that the hand information helps localize the target object.}
    \label{fig:further_output}
\end{figure}

    More testing examples in Figure~\ref{fig:further_output} were used to inspect the performances of the models, visually. The figure shows that our model estimated a more round shape of the center area, which is closer to the ground truth, than did the other models.  In particular, comparing the output of our method (\textit{HPLate}) with that of \textit{NoHand}, we found further evidence that the hand information is essential not only to localize an object of interest but also to decide which object would be of interest when multiple objects appear in the scenes.

\begin{figure}[!t]
    \centering
    \includegraphics[width=0.4\textwidth]{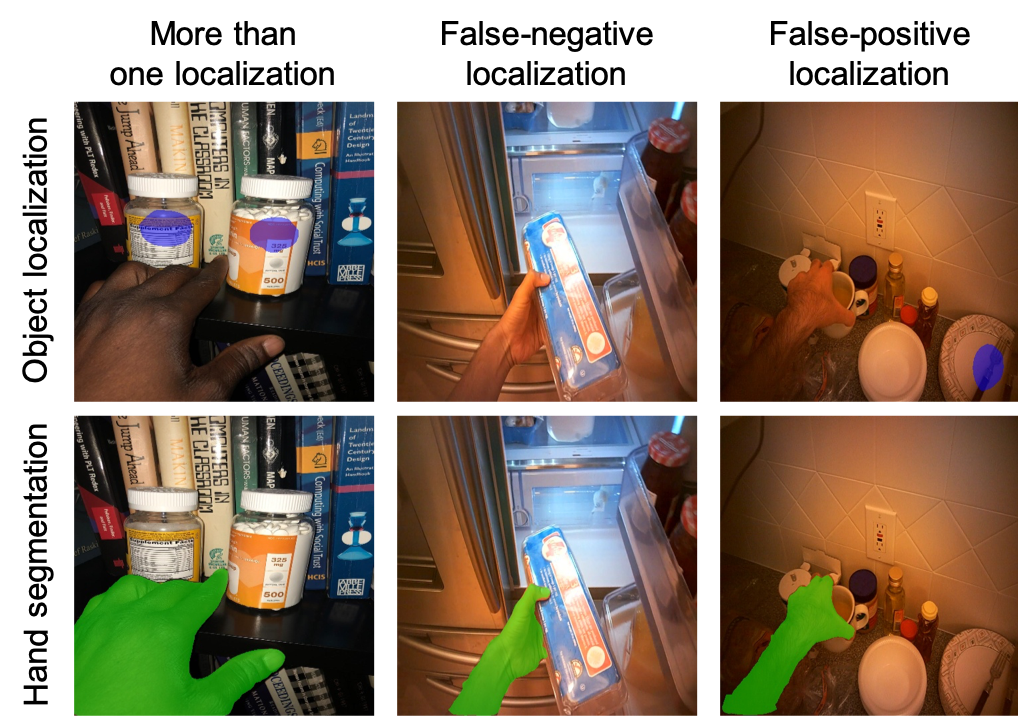}
    \caption{Failure cases of our method (\textit{HPLate}) despite the appropriate hand segmentation outputs on these testing examples.}
    \label{fig:neg_outputs}
\end{figure}

    \item[Failure cases:] Figure~\ref{fig:neg_outputs} shows failures cases of our model (\textit{HPLate}) in object localization. It can still be confused with two or more objects when the hand--object interaction is not so obvious. For example, the left example in the figure shows that our model was confused with the two objects in proximity of the hand. This confusion seems to be caused by an indiscernible interaction between the hand and the object in the example.  In addition, the middle example shows the false-negative localization of our model. Although our hand model successfully segmented the hand, the object model was unable to localize the object being held by the hand in the example. Furthermore, the false-positive output of our model is presented on the right of the figure. Despite the appropriate hand segmentation, our model localized a wrong object, which might be caused by the object occlusion by the hand. Further investigations are necessary to spot factors in these issues since the other methods also suffered from the issues in the same and/or different examples.

\begin{table}[!t]
    \centering
    \caption{Object localization performances of the methods on a different set of the TEgO data. For this analysis, the data from the blind person are separated from the data from the sighted person.}
    \resizebox{\columnwidth}{!}{
    \begin{tabular}{@{}llcccc@{}}
        \toprule
        Dataset & Model & $mIoU$ & $AP$ & $AP_{50}$ & $AP_{75}$ \\
        \midrule
        \multirow{8}{*}{\makecell{TEgO\\(blind)}} & NoHand & 0.71 & 0.52 & 0.90 & 0.54 \\
                              & Finetune & 0.71 & 0.51 & 0.90 & 0.52 \\
                              & MultiClass & 0.48 & 0.13 & 0.54 & 0.01 \\
                              & MultiClass-2x & \textbf{0.75} & \textbf{0.57} & 0.93 & 0.61 \\
                              & MultiTask & 0.31 & 0.03 & 0.17 & 0 \\
                              & MultiTask-2x & 0.71 & 0.50 & 0.91 & 0.50 \\
                              & \textit{HPAll} & 0.69 & 0.46 & 0.86 & 0.45 \\
                              & \textit{HPLate} & \textbf{0.75} & \textbf{0.57} & \textbf{0.94} & \textbf{0.64} \\
        \midrule
        \multirow{8}{*}{\makecell{TEgO\\(sighted)}} & NoHand & 0.74 & 0.54 & 0.93 & 0.58 \\
                              & Finetune & 0.74 & 0.55 & \textbf{0.96} & 0.57 \\
                              & MultiClass & 0.23 & 0.01 & 0.05 & 0 \\
                              & MultiClass-2x & 0.73 & 0.52 & 0.94 & 0.51 \\
                              & MultiTask & 0.52 & 0.17 & 0.63 & 0.03 \\
                              & MultiTask-2x & 0.74 & 0.53 & \textbf{0.96} & 0.54 \\
                              & \textit{HPAll} & 0.73 & 0.51 & 0.94 & 0.50 \\
                              & \textit{HPLate} & \textbf{0.76} & \textbf{0.59} & \textbf{0.96} & \textbf{0.64} \\
        \bottomrule
    \end{tabular}
    }
    \label{tab:tego_local_quant_analysis}
\end{table}

    \item[More in TEgO:] Data from blind people may not have some implicit information about the object of interest (\eg, placement of the target object at the center of the camera frame). Hence, in data from blind people, hands may be a reliable indication of a target object. As only TEgO includes data collected by a blind person, we evaluated all the methods with the TEgO data collected by the blind person. In Table~\ref{tab:tego_local_quant_analysis}, we report the performance of each model on data from sighted and blind people in the TEgO dataset, respectively.  For this analysis, the data from the sighted person and the data from the blind person were separately used to train and test each model. Our method, \textit{HPLate}, outperformed the other methods not only on TEgO (sighted) but also on TEgO (blind). In particular, compared to the baseline model (\textit{Finetune}), \textit{HPLate} achieves the performance improvement, on average, by 12\% and 7\% on the TEgO (blind) and TEgO (sighted) datasets, respectively.
    
\end{description}

\section{Discussion}

In this paper, our hand-primed object localization model showed the effectiveness of using hand information for object localization in egocentric vision.
We observed that our model worked well not only on the datasets from sighted people but also on the dataset that contains the hand--object interactions of the blind person.
We also saw its potential in being extended to applications in diverse domains, such as computer vision problems and assistive technologies.
In particular, assistive technologies that take egocentric input from users with visual impairments and employ computer vision models to understand the input may benefit from our approach when estimating a region of interest of the users and recognizing their object or activity.

Our approach has some limitations, which however are valuable guidance on our future directions.
First, a large dataset from the blind population may lead us to have more generalizable evaluation of the methods including our approach. Currently, assistive systems powered by state-of-art computer vision models suffer from a lack of datasets from this specific population~\cite{gurari2018vizwiz}.  As such assistive technologies can benefit from ample data, we are seeking more data collected by visually impaired people.

As our model only estimates the center area of an object of interest, an additional method is required to extract only the object of interest from the input image for further tasks, such as fine-grained object recognition. Prior work used the fixed size of a bounding box to extract the object of interest~\cite{ma2016going}, but such a naive approach may not work well on different sizes and shapes of objects. Object detection, such as region proposal network~\cite{ren2015faster}, may need to be incorporated to extract only a target object, but further research is required to use the contextual information (\ie hands) in such object detection.
Also, developing a hand detection model that detects an egocentric hand simply with a bounding box might be another cost-wise direction. Perhaps replacing the hand segmentation task with the hand detection task may lose some information about the hand, but it could be compensated by more labeled hand data. This hypothesis needs to be confirmed by further analysis. We leave this analysis as our future work.


Last, egocentric data collected with wearable cameras can include both left and right hands and two different objects being interacted with the left and right hands, respectively. In this case, it would be more natural and helpful to localize an object of interest for each hand, separately, to understand the egocentric context, more accurately.
In addition, prior work shows that video data could contain more information about understanding users' interactions with objects in the egocentric vision~\cite{ren2010figure, lee2012discovering, furnari2017next}.
Considering not only static images but also videos, we are currently improving our model to learn left and right hands and localize an object of interest for each hand.



\section{Conclusion}
We proposed an object localization model reinforced by hand information. In our approach, the output of the hand segmentation network is infused to the object localization network to prime the localization model to use the hand information for object localization in egocentric vision.
Our evaluation demonstrates the effectiveness of using the hand segmentation feedback for object localization --- estimating the center area of a target object.
It also shows that explicit infusion of the hand information into an object localization network achieves more precise localization than do the other approaches.
We believe that our method can be further employed in other applications that need to understand hand--object interactions, such as object/action recognition and assistive systems for people with visual impairments.

\section{Acknowledgments}
The authors thank the anonymous reviewers for their helpful comments on an earlier version of this work.  This work is supported by NIDILRR (\#90REGE0008).

{\small
\bibliographystyle{ieee}
\bibliography{main}

\begin{thebibliography}{10}\itemsep=-1pt

\bibitem{balata2015blindcamera}
J.~Balata, Z.~Mikovec, and L.~Neoproud.
\newblock Blindcamera: Central and golden-ratio composition for blind
  photographers.
\newblock In {\em Proceedings of the Mulitimedia, Interaction, Design and
  Innnovation}, page~8. ACM, 2015.

\bibitem{bambach2015lending}
S.~Bambach, S.~Lee, D.~J. Crandall, and C.~Yu.
\newblock Lending a hand: Detecting hands and recognizing activities in complex
  egocentric interactions.
\newblock In {\em Proceedings of the IEEE International Conference on Computer
  Vision}, pages 1949--1957, 2015.

\bibitem{bertasius2017unsupervised}
G.~Bertasius, H.~Soo~Park, S.~X. Yu, and J.~Shi.
\newblock Unsupervised learning of important objects from first-person videos.
\newblock In {\em Proceedings of the IEEE International Conference on Computer
  Vision}, pages 1956--1964, 2017.

\bibitem{biederman2017semantics}
I.~Biederman.
\newblock On the semantics of a glance at a scene.
\newblock In {\em Perceptual organization}, pages 213--253. Routledge, 2017.

\bibitem{bigham2010vizwizlocateit}
J.~P. Bigham, C.~Jayant, A.~Miller, B.~White, and T.~Yeh.
\newblock Vizwiz:: Locateit-enabling blind people to locate objects in their
  environment.
\newblock In {\em 2010 IEEE Computer Society Conference on Computer Vision and
  Pattern Recognition-Workshops}, pages 65--72. IEEE, 2010.

\bibitem{brady2013visual}
E.~Brady, M.~R. Morris, Y.~Zhong, S.~White, and J.~P. Bigham.
\newblock Visual challenges in the everyday lives of blind people.
\newblock In {\em Proceedings of the SIGCHI Conference on Human Factors in
  Computing Systems}, pages 2117--2126. ACM, 2013.

\bibitem{cai2016understanding}
M.~Cai, K.~M. Kitani, and Y.~Sato.
\newblock Understanding hand-object manipulation with grasp types and object
  attributes.
\newblock In {\em Robotics: Science and Systems}, volume~3, 2016.

\bibitem{Caruana1993MultitaskLA}
R.~Caruana.
\newblock Multitask learning: A knowledge-based source of inductive bias.
\newblock In {\em ICML}, 1993.

\bibitem{choi2017robust}
C.~Choi, S.~Ho~Yoon, C.-N. Chen, and K.~Ramani.
\newblock Robust hand pose estimation during the interaction with an unknown
  object.
\newblock In {\em Proceedings of the IEEE International Conference on Computer
  Vision}, pages 3123--3132, 2017.

\bibitem{damen2018scaling}
D.~Damen, H.~Doughty, G.~Maria~Farinella, S.~Fidler, A.~Furnari, E.~Kazakos,
  D.~Moltisanti, J.~Munro, T.~Perrett, W.~Price, et~al.
\newblock Scaling egocentric vision: The epic-kitchens dataset.
\newblock In {\em Proceedings of the European Conference on Computer Vision
  (ECCV)}, pages 720--736, 2018.

\bibitem{davis2006relationship}
J.~Davis and M.~Goadrich.
\newblock The relationship between precision-recall and roc curves.
\newblock In {\em Proceedings of the 23rd international conference on Machine
  learning}, pages 233--240. ACM, 2006.

\bibitem{deng2009imagenet}
J.~Deng, W.~Dong, R.~Socher, L.-J. Li, K.~Li, and L.~Fei-Fei.
\newblock Imagenet: A large-scale hierarchical image database.
\newblock In {\em Computer Vision and Pattern Recognition, 2009. CVPR 2009.
  IEEE Conference on}, pages 248--255. Ieee, 2009.

\bibitem{divvala2009empirical}
S.~K. Divvala, D.~Hoiem, J.~H. Hays, A.~A. Efros, and M.~Hebert.
\newblock An empirical study of context in object detection.
\newblock In {\em 2009 IEEE Conference on computer vision and Pattern
  Recognition}, pages 1271--1278. IEEE, 2009.

\bibitem{fathi2012learning}
A.~Fathi, Y.~Li, and J.~M. Rehg.
\newblock Learning to recognize daily actions using gaze.
\newblock In {\em European Conference on Computer Vision}, pages 314--327.
  Springer, 2012.

\bibitem{fathi2011learning}
A.~Fathi, X.~Ren, and J.~M. Rehg.
\newblock Learning to recognize objects in egocentric activities.
\newblock In {\em CVPR 2011}, pages 3281--3288. IEEE, 2011.

\bibitem{furnari2017next}
A.~Furnari, S.~Battiato, K.~Grauman, and G.~M. Farinella.
\newblock Next-active-object prediction from egocentric videos.
\newblock {\em Journal of Visual Communication and Image Representation},
  49:401 -- 411, 2017.

\bibitem{gosselin2009evidence}
N.~Gosselin-Kessiby, J.~F. Kalaska, and J.~Messier.
\newblock Evidence for a proprioception-based rapid on-line error correction
  mechanism for hand orientation during reaching movements in blind subjects.
\newblock {\em Journal of Neuroscience}, 29(11):3485--3496, 2009.

\bibitem{guo2018investigating}
A.~Guo, S.~McVea, X.~Wang, P.~Clary, K.~Goldman, Y.~Li, Y.~Zhong, and J.~P.
  Bigham.
\newblock Investigating cursor-based interactions to support non-visual
  exploration in the real world.
\newblock In {\em Proceedings of the 20th International ACM SIGACCESS
  Conference on Computers and Accessibility}, pages 3--14. ACM, 2018.

\bibitem{gurari2018vizwiz}
D.~Gurari, Q.~Li, A.~J. Stangl, A.~Guo, C.~Lin, K.~Grauman, J.~Luo, and J.~P.
  Bigham.
\newblock Vizwiz grand challenge: Answering visual questions from blind people.
\newblock {\em arXiv preprint arXiv:1802.08218}, 2018.

\bibitem{he2017mask}
K.~He, G.~Gkioxari, P.~Doll{\'a}r, and R.~Girshick.
\newblock Mask r-cnn.
\newblock In {\em Computer Vision (ICCV), 2017 IEEE International Conference
  on}, pages 2980--2988. IEEE, 2017.

\bibitem{huang2018predicting}
Y.~Huang, M.~Cai, Z.~Li, and Y.~Sato.
\newblock Predicting gaze in egocentric video by learning task-dependent
  attention transition.
\newblock In {\em Proceedings of the European Conference on Computer Vision
  (ECCV)}, pages 754--769, 2018.

\bibitem{jayant2011supporting}
C.~Jayant, H.~Ji, S.~White, and J.~P. Bigham.
\newblock Supporting blind photography.
\newblock In {\em The proceedings of the 13th international ACM SIGACCESS
  conference on Computers and accessibility}, pages 203--210. ACM, 2011.

\bibitem{kacorri2017people}
H.~Kacorri, K.~M. Kitani, J.~P. Bigham, and C.~Asakawa.
\newblock People with visual impairment training personal object recognizers:
  Feasibility and challenges.
\newblock In {\em Proceedings of the 2017 CHI Conference on Human Factors in
  Computing Systems}, pages 5839--5849. ACM, 2017.

\bibitem{kingma2014adam}
D.~P. Kingma and J.~Ba.
\newblock Adam: A method for stochastic optimization.
\newblock {\em arXiv preprint arXiv:1412.6980}, 2014.

\bibitem{lee2019revisiting}
H.~J. P. S. J.~E. Lee, Kyungjun and H.~Kacorri.
\newblock Revisiting blind photography in the context of teachable object
  recognizers.
\newblock In {\em Proceedings of the 21st International ACM SIGACCESS
  Conference on Computers and Accessibility}, ASSETS '19, New York, NY, USA,
  2019. ACM.

\bibitem{lee2019hands}
K.~Lee and H.~Kacorri.
\newblock Hands holding clues for object recognition in teachable machines.
\newblock In {\em Proceedings of the 2019 CHI Conference on Human Factors in
  Computing Systems}. ACM, 2019.

\bibitem{lee2012discovering}
Y.~J. {Lee}, J.~{Ghosh}, and K.~{Grauman}.
\newblock Discovering important people and objects for egocentric video
  summarization.
\newblock In {\em 2012 IEEE Conference on Computer Vision and Pattern
  Recognition}, pages 1346--1353, June 2012.

\bibitem{li2013learning}
Y.~Li, A.~Fathi, and J.~M. Rehg.
\newblock Learning to predict gaze in egocentric video.
\newblock In {\em Proceedings of the IEEE International Conference on Computer
  Vision}, pages 3216--3223, 2013.

\bibitem{li2018eye}
Y.~Li, M.~Liu, and J.~M. Rehg.
\newblock In the eye of beholder: Joint learning of gaze and actions in first
  person video.
\newblock In {\em Proceedings of the European Conference on Computer Vision
  (ECCV)}, pages 619--635, 2018.

\bibitem{li2017fully}
Y.~Li, H.~Qi, J.~Dai, X.~Ji, and Y.~Wei.
\newblock Fully convolutional instance-aware semantic segmentation.
\newblock In {\em Proceedings of the IEEE Conference on Computer Vision and
  Pattern Recognition}, pages 2359--2367, 2017.

\bibitem{li2015delving}
Y.~Li, Z.~Ye, and J.~M. Rehg.
\newblock Delving into egocentric actions.
\newblock In {\em Proceedings of the IEEE Conference on Computer Vision and
  Pattern Recognition}, pages 287--295, 2015.

\bibitem{long2015fully}
J.~Long, E.~Shelhamer, and T.~Darrell.
\newblock Fully convolutional networks for semantic segmentation.
\newblock In {\em Proceedings of the IEEE conference on computer vision and
  pattern recognition}, pages 3431--3440, 2015.

\bibitem{ma2016going}
M.~Ma, H.~Fan, and K.~M. Kitani.
\newblock Going deeper into first-person activity recognition.
\newblock In {\em Proceedings of the IEEE Conference on Computer Vision and
  Pattern Recognition}, pages 1894--1903, 2016.

\bibitem{meng2014neural}
Y.~Meng, X.~Ye, and B.~D. Gonsalves.
\newblock Neural processing of recollection, familiarity and priming at
  encoding: Evidence from a forced-choice recognition paradigm.
\newblock {\em Brain research}, 1585:72--82, 2014.

\bibitem{mottaghi2014role}
R.~Mottaghi, X.~Chen, X.~Liu, N.-G. Cho, S.-W. Lee, S.~Fidler, R.~Urtasun, and
  A.~Yuille.
\newblock The role of context for object detection and semantic segmentation in
  the wild.
\newblock In {\em Proceedings of the IEEE Conference on Computer Vision and
  Pattern Recognition}, pages 891--898, 2014.

\bibitem{murphy2004using}
K.~P. Murphy, A.~Torralba, and W.~T. Freeman.
\newblock Using the forest to see the trees: A graphical model relating
  features, objects, and scenes.
\newblock In {\em Advances in neural information processing systems}, pages
  1499--1506, 2004.

\bibitem{oliva2007role}
A.~Oliva and A.~Torralba.
\newblock The role of context in object recognition.
\newblock {\em Trends in cognitive sciences}, 11(12):520--527, 2007.

\bibitem{palmer1975effects}
t.~E. Palmer.
\newblock The effects of contextual scenes on the identification of objects.
\newblock {\em Memory \& Cognition}, 3:519--526, 1975.

\bibitem{pfister2015flowing}
T.~Pfister, J.~Charles, and A.~Zisserman.
\newblock Flowing convnets for human pose estimation in videos.
\newblock In {\em Proceedings of the IEEE International Conference on Computer
  Vision}, pages 1913--1921, 2015.

\bibitem{pirsiavash2012detecting}
H.~Pirsiavash and D.~Ramanan.
\newblock Detecting activities of daily living in first-person camera views.
\newblock In {\em 2012 IEEE Conference on Computer Vision and Pattern
  Recognition}, pages 2847--2854. IEEE, 2012.

\bibitem{proske2012proprioceptive}
U.~Proske and S.~C. Gandevia.
\newblock The proprioceptive senses: their roles in signaling body shape, body
  position and movement, and muscle force.
\newblock {\em Physiological reviews}, 92(4):1651--1697, 2012.

\bibitem{rabinovich2007objects}
A.~Rabinovich, A.~Vedaldi, C.~Galleguillos, E.~Wiewiora, and S.~J. Belongie.
\newblock Objects in context.
\newblock In {\em ICCV}, volume~1, page~5. Citeseer, 2007.

\bibitem{ren2015faster}
S.~Ren, K.~He, R.~Girshick, and J.~Sun.
\newblock Faster r-cnn: Towards real-time object detection with region proposal
  networks.
\newblock In {\em Advances in neural information processing systems}, pages
  91--99, 2015.

\bibitem{ren2010figure}
X.~{Ren} and C.~{Gu}.
\newblock Figure-ground segmentation improves handled object recognition in
  egocentric video.
\newblock In {\em 2010 IEEE Computer Society Conference on Computer Vision and
  Pattern Recognition}, pages 3137--3144, June 2010.

\bibitem{rogez20143d}
G.~Rogez, M.~Khademi, J.~Supan{\v{c}}i{\v{c}}~III, J.~M.~M. Montiel, and
  D.~Ramanan.
\newblock 3d hand pose detection in egocentric rgb-d images.
\newblock In {\em European Conference on Computer Vision}, pages 356--371.
  Springer, 2014.

\bibitem{sherrington1907proprio}
C.~S. SHERRINGTON.
\newblock On the proprio-ceptive system, especially in its reflex aspect.
\newblock {\em Brain}, 29(4):467--482, 1907.

\bibitem{shrivastava2016contextual}
A.~Shrivastava and A.~Gupta.
\newblock Contextual priming and feedback for faster r-cnn.
\newblock In {\em European Conference on Computer Vision}, pages 330--348.
  Springer, 2016.

\bibitem{simonyan2014very}
K.~Simonyan and A.~Zisserman.
\newblock Very deep convolutional networks for large-scale image recognition.
\newblock {\em arXiv preprint arXiv:1409.1556}, 2014.

\bibitem{surie2007activity}
D.~Surie, T.~Pederson, F.~Lagriffoul, L.-E. Janlert, and D.~Sj{\"o}lie.
\newblock Activity recognition using an egocentric perspective of everyday
  objects.
\newblock In {\em International Conference on Ubiquitous Intelligence and
  Computing}, pages 246--257. Springer, 2007.

\bibitem{tavakoli2019digging}
H.~R. Tavakoli, E.~Rahtu, J.~Kannala, and A.~Borji.
\newblock Digging deeper into egocentric gaze prediction.
\newblock In {\em 2019 IEEE Winter Conference on Applications of Computer
  Vision (WACV)}, pages 273--282. IEEE, 2019.

\bibitem{tekin2019h+}
B.~Tekin, F.~Bogo, and M.~Pollefeys.
\newblock H+ o: Unified egocentric recognition of 3d hand-object poses and
  interactions.
\newblock In {\em Proceedings of the IEEE Conference on Computer Vision and
  Pattern Recognition}, pages 4511--4520, 2019.

\bibitem{torralba2003context}
{Torralba}, {Murphy}, {Freeman}, and {Rubin}.
\newblock Context-based vision system for place and object recognition.
\newblock pages 273--280 vol.1, Oct 2003.

\bibitem{tulving1990priming}
E.~Tulving and D.~L. Schacter.
\newblock Priming and human memory systems.
\newblock {\em Science}, 247(4940):301--306, 1990.

\bibitem{vazquez2012helping}
M.~V{\'a}zquez and A.~Steinfeld.
\newblock Helping visually impaired users properly aim a camera.
\newblock In {\em Proceedings of the 14th international ACM SIGACCESS
  conference on Computers and accessibility}, pages 95--102. ACM, 2012.

\bibitem{vazquez2014assisted}
M.~V{\'a}zquez and A.~Steinfeld.
\newblock An assisted photography framework to help visually impaired users
  properly aim a camera.
\newblock {\em ACM Transactions on Computer-Human Interaction (TOCHI)},
  21(5):25, 2014.

\bibitem{white2010easysnap}
S.~White, H.~Ji, and J.~P. Bigham.
\newblock Easysnap: real-time audio feedback for blind photography.
\newblock In {\em Adjunct proceedings of the 23nd annual ACM symposium on User
  interface software and technology}, pages 409--410. ACM, 2010.

\bibitem{wig2005reductions}
G.~S. Wig, S.~T. Grafton, K.~E. Demos, and W.~M. Kelley.
\newblock Reductions in neural activity underlie behavioral components of
  repetition priming.
\newblock {\em Nature neuroscience}, 8(9):1228, 2005.

\bibitem{yao2012describing}
J.~{Yao}, S.~{Fidler}, and R.~{Urtasun}.
\newblock Describing the scene as a whole: Joint object detection, scene
  classification and semantic segmentation.
\newblock In {\em 2012 IEEE Conference on Computer Vision and Pattern
  Recognition}, pages 702--709, June 2012.

\bibitem{zeiler2014visualizing}
M.~D. Zeiler and R.~Fergus.
\newblock Visualizing and understanding convolutional networks.
\newblock In {\em European conference on computer vision}, pages 818--833.
  Springer, 2014.

\bibitem{zhao2018face}
Y.~Zhao, S.~Wu, L.~Reynolds, and S.~Azenkot.
\newblock A face recognition application for people with visual impairments:
  Understanding use beyond the lab.
\newblock In {\em Proceedings of the 2018 CHI Conference on Human Factors in
  Computing Systems}, page 215. ACM, 2018.

\bibitem{zhong2013real}
Y.~Zhong, P.~J. Garrigues, and J.~P. Bigham.
\newblock Real time object scanning using a mobile phone and cloud-based visual
  search engine.
\newblock In {\em Proceedings of the 15th International ACM SIGACCESS
  Conference on Computers and Accessibility}, page~20. ACM, 2013.

\bibitem{zhu2015segdeepm}
Y.~Zhu, R.~Urtasun, R.~Salakhutdinov, and S.~Fidler.
\newblock segdeepm: Exploiting segmentation and context in deep neural networks
  for object detection.
\newblock In {\em Proceedings of the IEEE Conference on Computer Vision and
  Pattern Recognition}, pages 4703--4711, 2015.

\bibitem{zuo2018gaze}
Z.~Zuo, L.~Yang, Y.~Peng, F.~Chao, and Y.~Qu.
\newblock Gaze-informed egocentric action recognition for memory aid systems.
\newblock {\em IEEE Access}, 6:12894--12904, 2018.

\end{thebibliography}
}

\end{document}